\documentclass[10pt,twocolumn,letterpaper]{article}

\usepackage{iccv}
\usepackage{times}
\usepackage{epsfig}
\usepackage{graphicx}
\usepackage{amsmath}
\usepackage{amssymb}
\usepackage{subfigure}
\usepackage{amsfonts}

\usepackage[breaklinks=true,bookmarks=false]{hyperref}

\iccvfinalcopy 

\def\iccvPaperID{****} 

\ificcvfinal\fi

\begin{document}

\title{FR-Net:A Light-weight FFT Residual Net For Gaze Estimation}


\author{
Tao Xu$^1$, Bo Wu$^1$, Ruilong Fan$^1$, Yun Zhou$^2$ and Di Huang$^3$\\
$^1$ School of Software, Northwestern Polytechnical University, Xi'an, China\\
$^2$ Faculty of Education, Shaanxi Normal University, Xi'an, China \\
$^3$ School of Computer Science and Engineering, Beihang University, Beijing, China\\
{\tt\small xutao@nwpu.edu.cn, \{wubo9826, 2019602431\}@mail.nwpu.edu.cn} \\ 
{\tt\small zhouyun@snnu.edu.cn, dhuang@buaa.edu.cn}
}

\maketitle
\ificcvfinal\thispagestyle{empty}\fi

\begin{abstract}

Gaze estimation is a crucial task in computer vision, however, existing methods suffer from high computational costs, which limit their practical deployment in resource-limited environments. In this paper, we propose a novel lightweight model, FR-Net, for accurate gaze angle estimation while significantly reducing computational complexity. FR-Net utilizes the Fast Fourier Transform (FFT) to extract gaze-relevant features in frequency domains while reducing the number of parameters. Additionally, we introduce a shortcut component that focuses on the spatial domain to further improve the accuracy of our model. Our experimental results demonstrate that our approach achieves substantially lower gaze error angles (3.86 on MPII and 4.51 on EYEDIAP) compared to state-of-the-art gaze estimation methods, while utilizing 17 times fewer parameters (0.67M) and only 12\% of FLOPs (0.22B). Furthermore, our method outperforms existing lightweight methods in terms of accuracy and efficiency for the gaze estimation task. These results suggest that our proposed approach has significant potential applications in areas such as human-computer interaction and driver assistance systems.
\end{abstract}

\section{Introduction}

Gaze, as one of the most prominent cues of human attention, plays a crucial role in nonverbal communication. It serves as a tool to measure and enhance human attention and interest, and its significance in understanding human behavior and mental states cannot be overstated. 
The estimation of gaze is a well-established technique that finds application in a diverse range of fields, including but not limited to saliency detection\cite{ferrari_salient_2018}, assisted driving\cite{martin_dynamics_2018}, and human-computer interaction\cite{masse_tracking_2018}.


The advancements in deep learning have led to a shift in gaze estimation from a model-based approach\cite{guestrin_general_2006,valenti_combining_2012} to an appearance-based approach\cite{zhang_its_2017,chen_appearance-based_2019,kellnhofer_gaze360_2019,cheng_coarse--fine_2020,cheng_gaze_2022} that relies solely on facial images. Based on capturing facial images using a webcam, the appearance-based approach yields a higher precision gaze direction, whose results are already comparable to a professional eye-tracking device. Thereby it enabling this technology uses in a broader range of applications beyond the constraints of experimental scenarios.

\begin{figure}[htbp]
\centering
\includegraphics[width=0.48\textwidth]{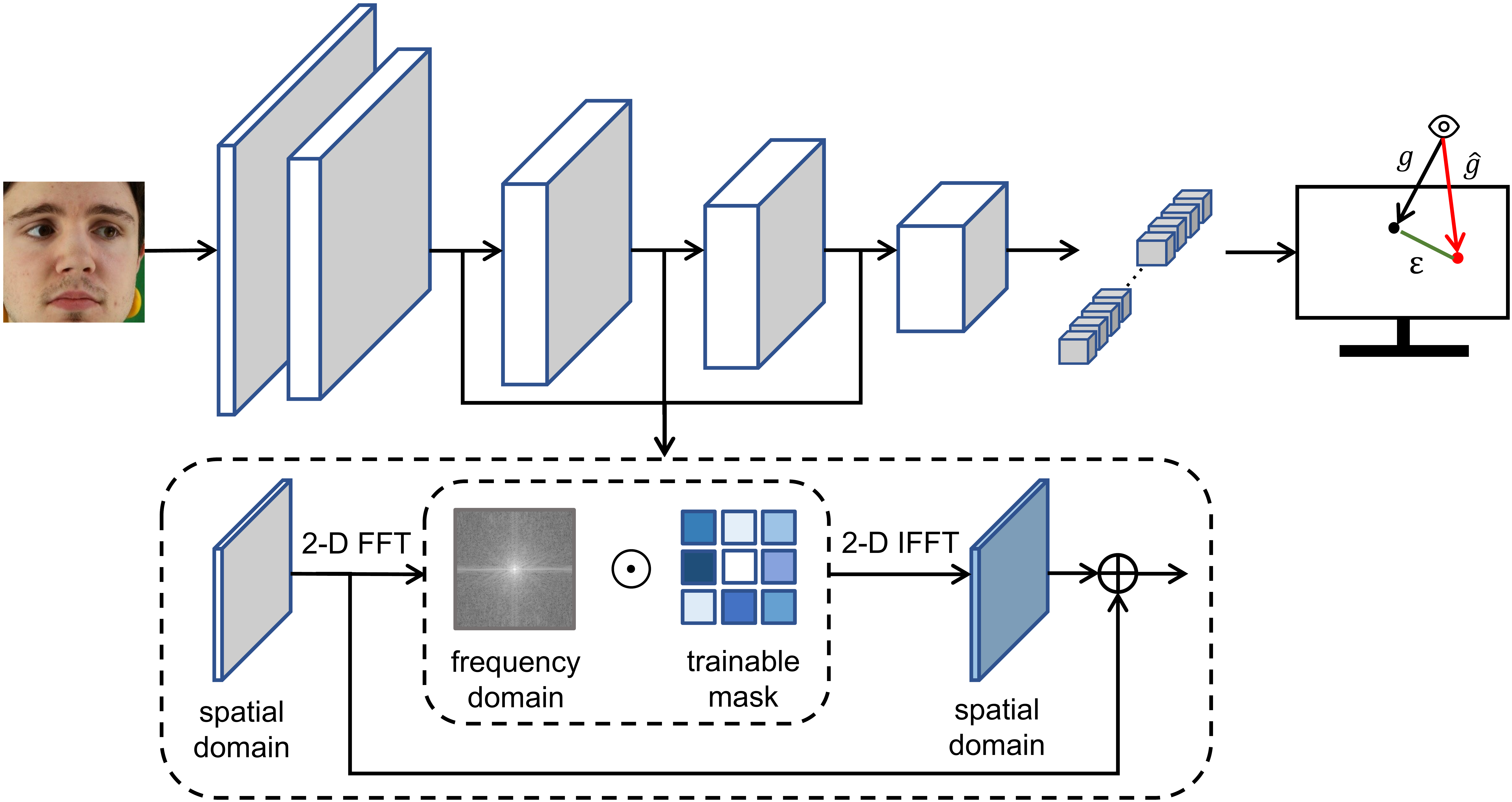}
\caption{Brief illustration of FR-Net. Core component: FFT Residual block is proposed to extract the pertinent frequency and spatial features related to eyes. Backbone Model: MobileVit v3}
\label{Fig:2 FR-Net briefGraph}
\end{figure}

\begin{figure}[h!]
\centering
\includegraphics[width=0.5\textwidth]{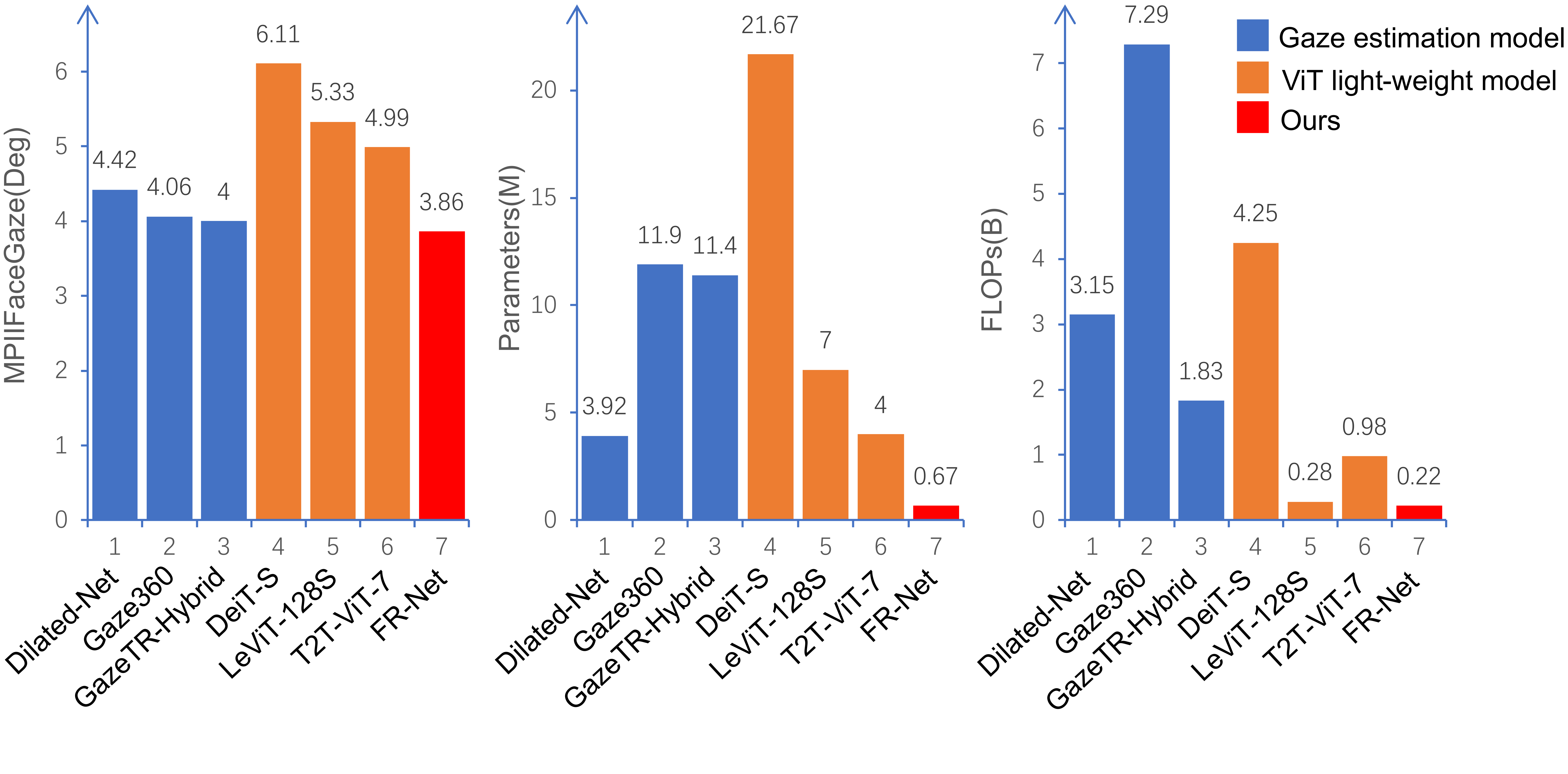}
\caption{Our model demonstrates substantial improvements over both state-of-the-art gaze estimation models and lightweight VIT algorithms. The figure showcases some of the results, validated on MPIIFaceGaze.}
\label{Fig:1 bar MPIIFaceGaze}
\end{figure}

The vision transformer (ViT) architecture dominates in most applications of the computer vision field, which uses self-attention mechanisms to process image patches in parallel. It also performs well on gaze estimation tasks, like \cite{cheng_gaze_2022,oh_self-attention_2022}, they combined the self-attention mechanism with CNN to investigate the effectiveness of gaze estimation and model generalization, and their method achieved the leading level at the time in some general gaze estimation datasets\cite{funes_mora_eyediap_2014,zhang_its_2017,kellnhofer_gaze360_2019}. However, a series of ViT models require a large number of parameters and heavy computation, making it hard to deploy on resource-constrained devices. It is a challenge to achieve good results with a small number of parameters.

Gaze estimation relies heavily on accurate localization of the eye region and the extraction of critical global features. Leveraging large receptive fields can facilitate the extraction of additional information from eye images. Current approaches commonly employ cropped face images from videos as inputs, with eye images cropped at a consistent high resolution of $W \times H$. Prior works, such as \cite{sugano_learning-by-synthesis_2014,zhang_mpiigaze_2019,ali_2020-deep_2020,chen_appearance-based_2019,krafka_eye_2016}, have reported using resolutions of $60 \times 36$, $60 \times 60$, $64 \times 96$, and $224 \times 224$, respectively.

Our work is motivated by three key desiderata for improving the performance of gaze estimation with minimal computational effort:
\begin{itemize}
    \item Firstly, to capture contextual information, large receptive fields are necessary. 
    \item Secondly, when a large kernel size is needed, FFT layers are more efficient than convolution layers \cite{chi_fast_2020}. 
    \item Finally, shortcut connections are vital, particularly for networks with very large kernels, to facilitate gradient flow and improve training \cite{ding_scaling_2022}
\end{itemize}

In this work, we present FR-Net, a novel lightweight model designed for efficient gaze estimation tasks. FR-Net introduces a Fast Fourier Transform (FFT) Residual Block and utilizes MobileViT v3 \cite{wadekar_mobilevitv3_2022} as its backbone architecture. The brief structure of FR-Net is shown in Figure \ref{Fig:2 FR-Net briefGraph}.  The proposed model aims to leverage the benefits of both frequency and spatial domain information. The FFT Residual Block applies FFT to the input feature map to extract frequency domain features and uses a global filter to capture latent features. It reduces the computational complexity from $O\left(N^{2}\right)$  to $O(N \log N)$. Moreover, the shortcut connection in the FFT Residual Block is designed to capture spatial domain information and improve model's performance. 
 
To evaluate the performance of our proposed method, we conduct experiments on publicly available databases: MPIIFaceGaze\cite{zhang_its_2017} and EYEDIAP\cite{funes_mora_eyediap_2014}. Our method achieves superior performance compared to existing methods while utilizing only a fraction of the parameters and computational resources required by prior methods. Since we adopt ViT as the backbone of our method and compare it with ViT's lightweight work on the gaze estimation task. Our approach outperforms ViT's lightweight work in both accuracy and parameter efficiency. Partial results in terms of angle error, parameters, and FLOPs are shown in Figure \ref{Fig:1 bar MPIIFaceGaze}.  Furthermore, our algorithm accurately localizes the eye region, as intended by the design, as evidenced by visual analysis.

In this work, we present our primary contributions as: 
\begin{enumerate}
    \item A novel model that achieves high accuracy in gaze estimation by utilizing a trainable FFT Residual Block, which effectively preserves gaze-relevant features in both the frequency and spatial domains.
    \item A low computational complexity version of the MobileViT architecture that incorporates the FFT Residual Block to reduce the number of parameters and operations while maintaining precision. This makes our proposed model suitable for deployment in resource-limited settings, expanding its practical potential.
\end{enumerate}

The article is organized as follows. Section 2 provides a comprehensive review of pertinent literature pertaining to gaze estimation and lightweight networks. Section 3 presents the detailed architecture of the proposed FR-Net. Section 4 conducts a comprehensive evaluation of the proposed approach via experiments on publicly available datasets, MPIIFaceGaze and EYEDIAP, as well as ablation experiments to examine the individual contributions of network components. Lastly, Section 5 and 6 draw conclusions and discuss future directions for research.

\section{Related Work} 
\subsection{Gaze Estimation}
The field of gaze estimation has conventionally been categorized into two primary approaches: model-based and appearance-based. Model-based gaze estimation employs eye structure and characteristics, including corneal reflection \cite{guestrin_general_2006} and pupil center \cite{valenti_combining_2012}, to estimate gaze. While it provides greater accuracy, it is restricted by outdoor conditions and necessitates specialized hardware.

The appearance-based gaze estimation methods leverages simple webcam-based gaze estimation and has gained popularity due to its ease of use. The advancement of gaze estimation has been facilitated by the use of datasets\cite{funes_mora_eyediap_2014, zhang_its_2017, kellnhofer_gaze360_2019,zhang_eth-xgaze_2020} that simulate real-world environments and take into account various factors such as light intensity, glasses, human species, and head posture. 

\begin{figure*}[ht]
\centering
\includegraphics[width=1\textwidth]{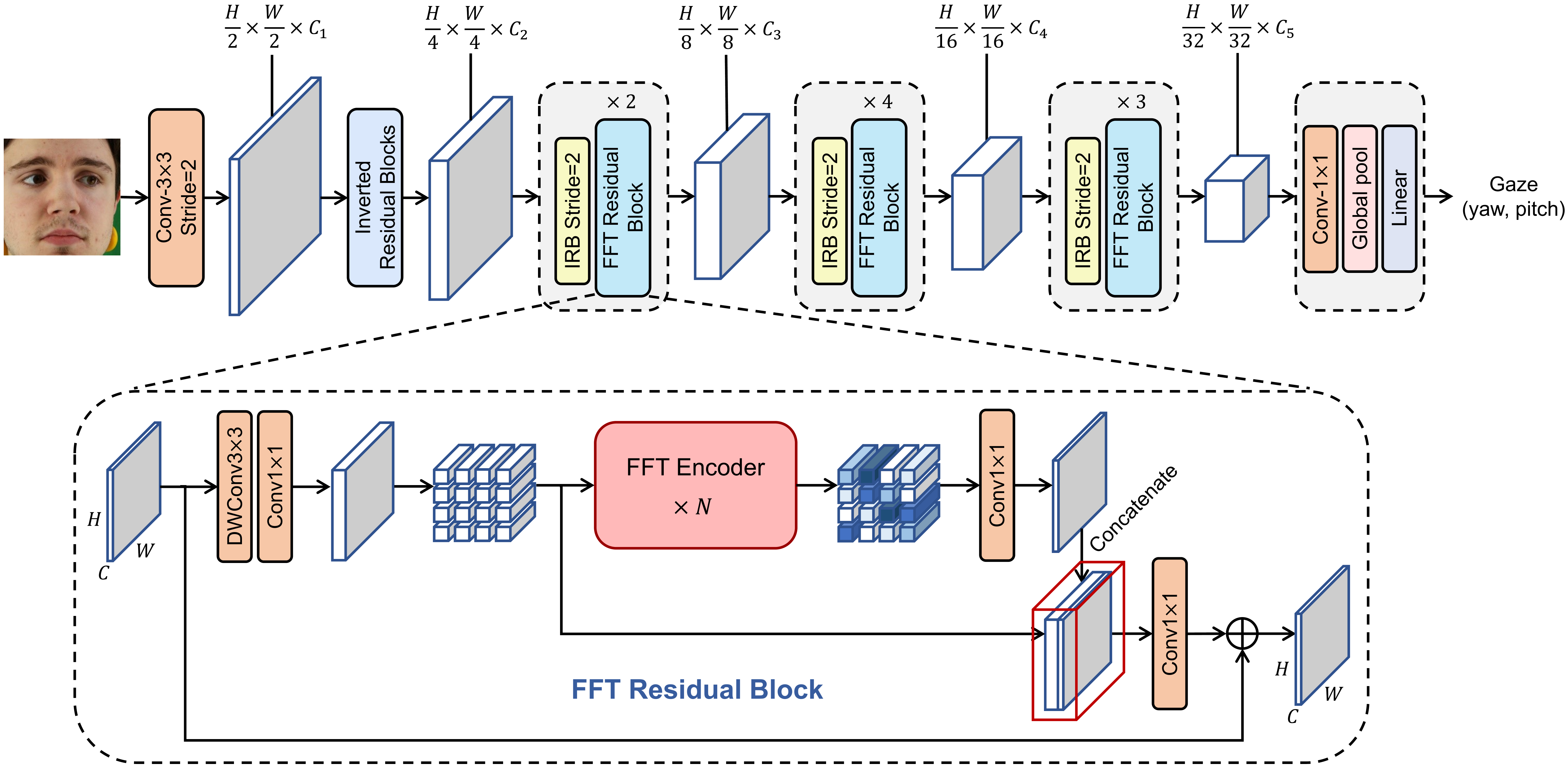}
\caption{The three FFT Residual Blocks are utilized to extract both depth spatial domain and frequency domain features, which are then fused to form the input for the subsequent layer. The input dimension of the FFT Encoder is 64, 80, and 96 in consecutive order. The $1\times1$ convolutional layer is employed to control the variation in the dimension of the feature map. The channels $C_1, C_2, C_3, C_4, C_5$ have dimensions of 16, 24, 48, 64, and 80, respectively. The final $1\times1$ convolutional layer has an output channel of 320, resulting in a two-dimensional output, representing yaw and pitch.}
\label{Fig:3 FRNet detail}
\end{figure*}

Gaze estimation has become a focal point in the field of human-computer interaction, with an increasing shift towards practical applications from laboratory research. Recent works have leveraged deep learning techniques. Zhang et al. \cite{zhang_its_2017} introduced a spatial weighting mechanism to encode the significance of the entire facial image in their approach. Meanwhile, Chen et al. \cite{chen_appearance-based_2019} utilized a dilated convolutional approach with three channels of data, consisting of left and right eye images and facial images, as input to their model. Kellnhofer et al. \cite{kellnhofer_gaze360_2019} acknowledged the importance of temporal order in gaze estimation and employed a combination of Convolutional Neural Networks (CNNs) and Long Short-Term Memory (LSTM) to infer gaze from continuous facial images. Cheng et al. \cite{cheng_gaze_2022} achieved state-of-the-art results on publicly available datasets by incorporating a Transformer architecture into their CNN-based model.

Generic gaze estimation techniques often exhibit large biases that can vary among individuals, posing a domain generalization problem. To address this issue, Yu et al. \cite{yu_improving_2019} synthesized gaze-reoriented images from reference samples and performed self-supervised domain adaptation for gaze reorientation. Cheng et al. \cite{cheng_puregaze_2022} proposed a self-adversarial framework for eliminating gaze-independent features in human faces, without requiring target samples. Liu et al. \cite{liu_generalizing_2021} adopted a plug-and-play cross-domain framework based on outlier-guided collaborative learning. Wang et al. \cite{wang_contrastive_2022} employed contrastive learning for the gaze estimation regression task, aiming to separate gaze-related features from gaze-unrelated features.

The present endeavors concentrate on enhancing model accuracy and generalization ability. With the advent of Vision Transformer (ViT) in this domain, models are becoming progressively larger, necessitating abundant computational resources, and posing challenges in resource-limited devices.

\subsection{Light-weight Net}
Lightweight Convolutional Neural Networks (CNNs) play a crucial role in the field of lightweight networks across diverse domains. The deep separable convolutional layer proposed by Xception\cite{chollet_xception_2017} effectively reduces parameters, which is a crucial factor for certain network architectures\cite{howard_mobilenets_2017,sandler_mobilenetv2_2018,zhang_shufflenet_2018}.  Meanwhile, MobileNet\cite{howard_mobilenets_2017} abandons the conventional methods of shrinking, pruning, quantizing, or compressing small models and instead incorporates the use of deep separable convolutional layers to make the network architecture more efficient. The introduction of backward residuals with linear bottlenecks in MobileNetV2\cite{sandler_mobilenetv2_2018} further optimizes the MobileNet architecture, leading to improved accuracy and reduced computation. 

Compared to other architectures, ViT has the advantage of adaptive weighting of inputs and global processing, demonstrating high performance across multiple domains. However, the computational cost of ViT limits its practicality in mobile applications. LeViT network \cite{graham_levit_2021} integrates transformer architecture into convolutional networks, resulting in improved efficiency and performance.
The MobileViT series of networks \cite{mehta_mobilevit_2022, mehta_separable_2022, wadekar_mobilevitv3_2022} combine the benefits of CNNs and Transformers, exhibiting outstanding performance on numerous mobile vision tasks. Despite having fewer parameters compared to other lightweight networks, the presence of a self-attentive mechanism in the network structure makes MobileViT series networks have higher FLOPs, higher constraints on device resources, and inadequate adherence to low latency requirements.

\section{Methods}
Appearance-based gaze estimation is a method of predicting the direction of a person's gaze by analyzing the appearance of their eyes or face. The basic idea behind appearance-based gaze estimation is that certain features of the eyes and face change depending on where a person is looking.

Our approach introduces the main idea illustrated in Figure \ref{Fig:2 FR-Net briefGraph}. It is motivated that leveraging large receptive fields can facilitate the extraction of more salient features from eye images, and FFT layers offer a much more computationally efficient alternative to convolutional layers when dealing with large kernel sizes. 

The model utilizes FFT instead of convolution to extract frequency domain features and reduce computation complexity.  Our proposed model offers a tradeoff between accuracy and efficiency by utilizing FFT-based techniques. 

Additionally, we design a series of shortcut components that selectively aggregates features from the spatial domain, effectively improving the model's accuracy. Our proposed model offers a tradeoff between accuracy and efficiency by utilizing FFT-based techniques.

\subsection{FFT Residual Block}
\label{section_FRB}
Our more detailed framework is shown in Figure \ref{Fig:3 FRNet detail}. the backbone of our model is MobileVit V3\cite{wadekar_mobilevitv3_2022}. We have kept the main structure and the lightweight unit: inverted residual block and proposed FFT Residual Block replacing transformer encoder, to improve model accuracy while reducing computational complexity.

The inverted residual block of two types of convolutions operations, namely depth-separable convolution and point-wise convolution. The depth-separable convolution performs separate convolutions on the input feature map’s channels, which significantly reduces the number of parameters, but it separates the channel and spatial information and cannot utilize feature information at the same spatial location from different channels. 

The self-attention mechanism in the transformer architecture can Effectively extract intrinsic features after the inverted residual block. It is the computationally intensive part, resulting in a significant increase in FLOPs, It presents a formidable challenge to the limited computing resources of mobile devices and hinders their capability for low-latency and real-time processing. To this end, we present the FFT Residual Block as a solution for the MobileViT network series. The FFT Residual Block is designed to be lightweight, and it effectively extracts gaze-related features from both the spatial and frequency domains by using a simple fusion method. The key component of the FFT Residual Block is the FFT Encoder, which is depicted in Figure \ref{Fig:4 FFT Encoder}.  There are two main parts: 

\begin{figure}[htbp]
\centering
\includegraphics[width=0.22 \textwidth]{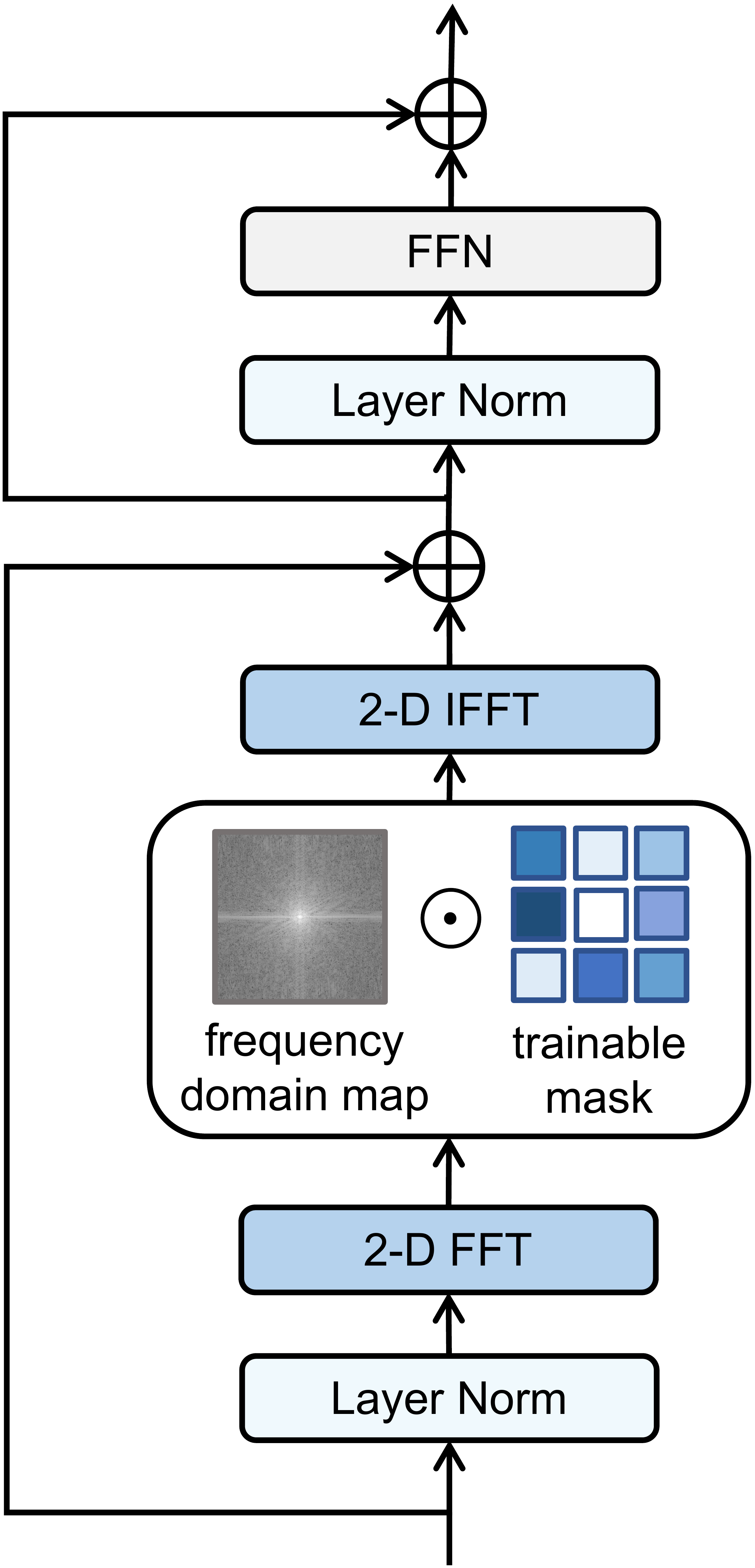}
\caption{The structure of FFT Encoder: Our proposed approach involves replacing the self-attention mechanism utilized in the Transformer Encoder architecture with FFT Encoder in the frequency domain. This modification efficiently extracts gaze-related features in the frequency domain.}
\label{Fig:4 FFT Encoder}
\end{figure}

\subsubsection{Calculating Convolutions Using the FFT}


FFT offers an efficient way to perform convolution by exploiting the convolution theorem, which states that convolution in the time domain is equivalent to multiplication in the frequency domain. By applying FFT to the inputs and kernels, multiplying their corresponding frequency components, and then applying the inverse FFT to the result, we can obtain the convolution in the time domain.

\begin{figure}[htbp]
\centering
\includegraphics[width=0.45 \textwidth]{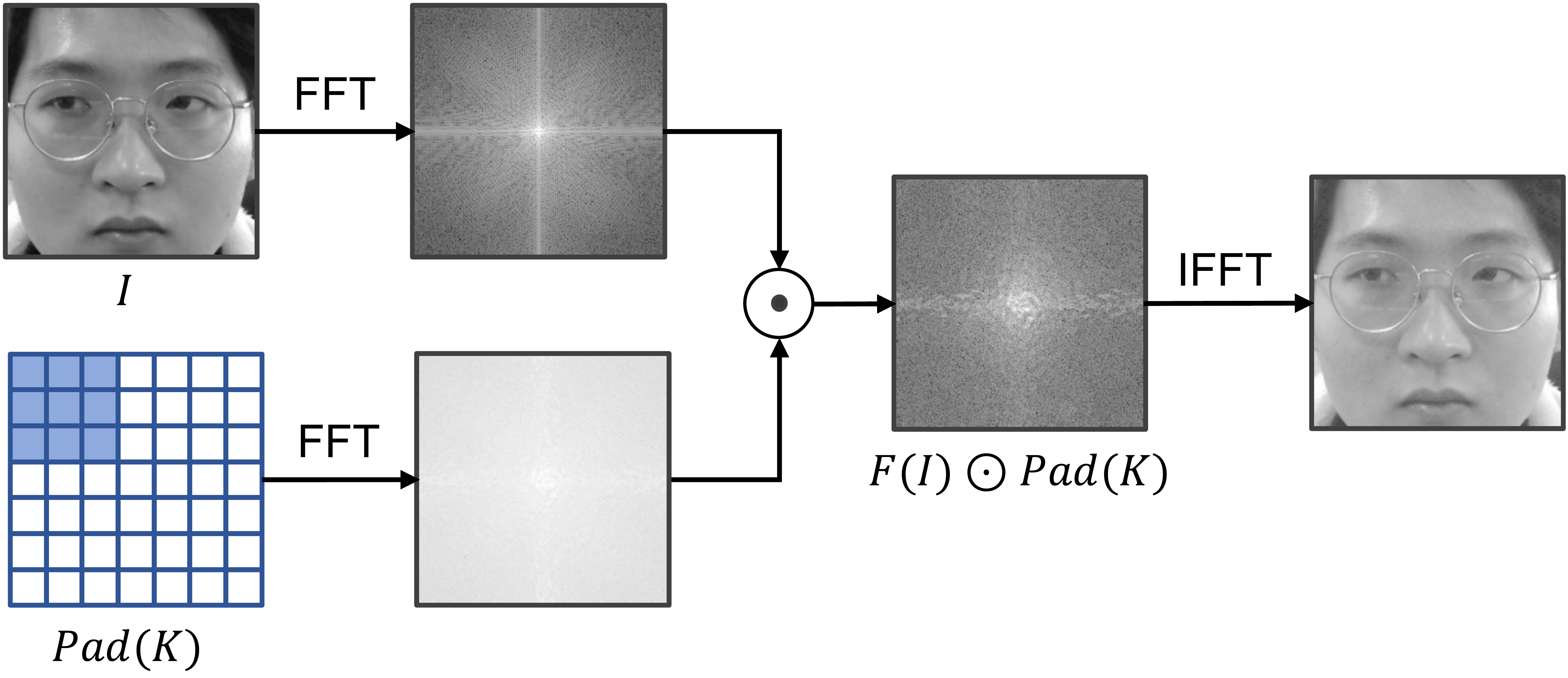}
\caption{Calculating convolutions using FFT}
\label{Fig:5 FFT PadK}
\end{figure}

Here are the general steps to replace convolution with FFT, shown in Figure \ref{Fig:5 FFT PadK}:
\begin{enumerate}
\item Pad the kernel: First, we need to make sure that the kernel has the same size as the input image. If they are not of the same size, we can pad the smaller kernels with zeros to make them the same length.
\item Compute the FFT of the inputs: Use the FFT algorithm to compute the DFT of images and kernels.
\item Multiply the Fourier transforms of the input signals: Multiply the Fourier transforms of the two input signals element-wise to obtain the Fourier transform of their convolution
\item Compute the inverse FFT: Use the inverse FFT algorithm to compute the inverse DFT of the product obtained in Step 3.
\item Truncate the result: The result of the inverse FFT will be a sequence of complex numbers. We can truncate the imaginary part of the result and take only the real part to obtain the result of the convolution operation.
\end{enumerate}



A single convolution can be represented as an equation:
\begin{equation}
    Y = X \ast K
\end{equation}

where $\ast$ is a convolution operation, X, F represents the input image and the filter.

We can rewrite equation (1) in the Fourier domain as equation (2)
\begin{equation}
y = F(X) \cdot F(K) = x \cdot k
\end{equation}

where $\cdot$ represents a Hadamard product. Hadamard product is an element-wise operation. It is required to pad the filter $f$ to let the size equal to $x$.

Then take the inverse Fourier transform of the multiplication result: $y$. Now the result of the inverse Fourier transform would be similar to the output of the convolution. 

the two-dimensional discrete Fourier transform is represented as follows:
\begin{equation}
    F(x, y)=\sum_{m=0}^{M-1} \sum_{n=0}^{N-1} f(m,n) e^{-j 2 \pi\left(\frac{u x}{M}+\frac{v y}{N}\right)}  \label{2ddft}
\end{equation}
where x, y represents the input of 2D signal and N, M presents the dimension of the input.

FFT represents the fast Fourier transform. The fast Fourier transform (FFT) algorithm effectively reduces the computational complexity from $O\left(N^{2}\right)$  to $O(N \log N)$ by exploiting the periodic and symmetric nature of $W_{N}$, thereby simplifying the DFT calculation for digital computers. The inverse fast Fourier transform (IFFT) serves a similar purpose to the inverse discrete Fourier transform (IDFT) and enables efficient computation.

To further reduce the computational effort, we get rid of the frequency domain transformation of the kernel and instead adopt a global trainable mask that matches the input dimensions. This approach draws inspiration from the methodology presented in \cite{rao_global_2021}. It reduces the process of making FFT on the kernel. In the back-propagation optimization process, the two are equivalent, and the mask is equivalent to:

\begin{equation}
Mask = FFT(padding(F))
\label{2ddft}
\end{equation}

This trainable mask can also be seen as a large convolution to extract significant features in the frequency domain, but it is significantly faster than traditional convolution for large kernel sizes.

\begin{figure}[!h]
\centering
\includegraphics[width=0.45 \textwidth]{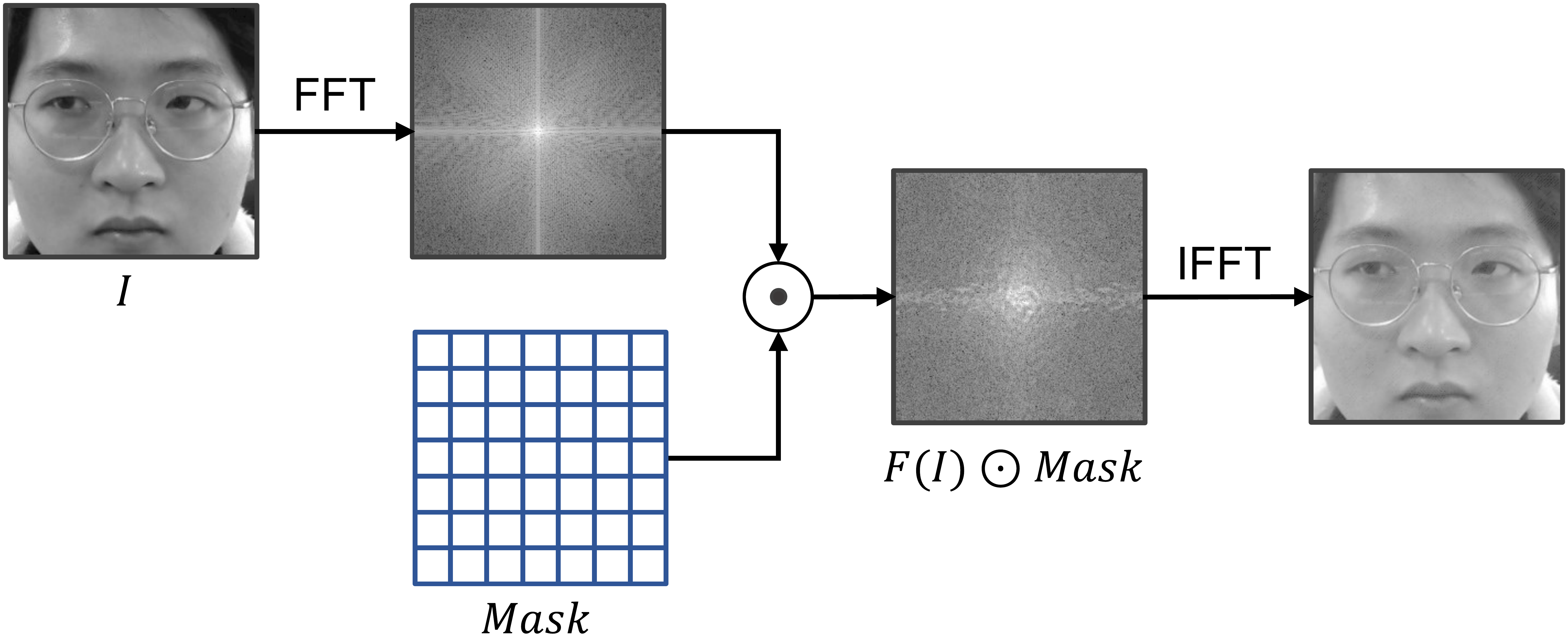}
\caption{The main idea applies FFT in our model.}
\label{Fig:6 FFT Mask}
\end{figure}



\subsection{Spatial feature extraction}
The fast Fourier transform (FFT) is known to capture only frequency domain features and lacks spatial location information. Thus, it is crucial to also extract original features with significant spatial locations in the spatial domain. 


The FFT convolution components endeavor to extract frequency features in a global manner, which can be viewed as a very large kernel. In \cite{ding_scaling_2022}, Ding et.al prove that incorporating a shortcut connection is vital to enhance the performance of networks employing large kernels.


In light of the aforementioned points, our approach incorporates a series of shortcut connections to improve the ability of our model to extract spatial features. The first one is added after the convolution

\begin{equation}
    y=F\left(x,\left\{W_{i}\right\}\right)+x
\end{equation}

 where $x$ and $y$  are the input and output, and $F\left(x,\left\{W_{i}\right\}\right)$ represents the FFT mapping to be learned.

Notably, Another shortcut connection is added after the group of FFT Encoders. We make a modification traditional way. It concatenated the layer extracting frequency-domain features and spatial information rather than directly element-wise adding to the FFT-extracted features. 

\begin{equation}
     F_{fusion}(x) = Concat(FFT_{encoder}(x,W),x)
     \label{}
\end{equation}
where $x$ and $F_{fusion}(x)$ represent the input and output of core components of FFT Residual Block, The function $FFT_{encoder}(x,W)$ represents the frequency feature mapping to be learned by a group of FFT Encoders. An ablation study confirms the importance of this design in our model.

\subsection{Implementation Details}
 In this study, the network model was implemented using the PyTorch framework, which utilized a smoothed L1 loss function and the AdamW optimizer throughout the training process. The resolution size of the images was set at $256\times256$. The network was initialized with pre-training parameters from the ETH-XGaze dataset for the MPIIFaceGaze and EYEDIAP datasets, with the initial learning rate set at 0.0004 and decaying to 0.00004 after 10 epochs. The batch size used for the assessment of MPIIFaceGaze was 64. We conducted our experiments on a computing system equipped with an Intel(R) Core(TM) i7-11800H CPU, NVIDIA RTX 3090 GPU, and 32G RAM.

\section{Experiment}


We conducted a thorough evaluation of the proposed model using a widely employed and publicly accessible dataset in the field of gaze estimation. 
Our work is compared with state of art works on gaze estimation and Vit lightweight work using performance metrics such as angular error, parameters, FLOPs, and inference time. To validate our design choices, we present a visual analysis of the essential components of our model. Furthermore, we perform ablation experiments to investigate the contributions of key components in our approach.

\subsection{Experimental Settings}


\textbf{Dataset for evaluation}
To evaluate the performance of the model, we utilized two publicly accessible gaze estimation datasets, namely MPIIFaceGaze\cite{zhang_its_2017} and EYEDIAP\cite{funes_mora_eyediap_2014}. The MPIIFaceGaze dataset was collected through a laptop webcam and comprises 213,659 images from 15 subjects, with 3000 samples per subject for evaluation. The EYEDIAP dataset is comprised of 237 minutes of video captured via an HD webcam from 16 subjects, with screen targets and swinging blobs serving as visual stimuli. The data for MPIIFaceGaze and EYEDIAP was normalized according to Cheng's review\cite{cheng_appearance-based_2021}, with the MPIIFaceGaze evaluated via the leave-one-out approach and the EYEDIAP evaluated using 4-fold cross-validation.

\textbf{Evaluation Metric}
For the gaze estimation task, we utilize the angular error as the primary metric, which quantifies the deviation between the estimated and ground truth gaze directions. A lower angular error indicates better accuracy and superior performance.

\begin{equation}
    \epsilon = \frac{g \cdot \hat{g}}{\Vert g \Vert \Vert \hat{g} \Vert}
\end{equation}

where $\epsilon$ is the angular error. $g \in \mathbb{R}^{3}$ and $\hat{g} \in \mathbb{R}^{3}$ are the true and estimated gaze directions, respectively.

To assess the effectiveness of our model, we evaluate it using multiple metrics, including the number of model parameters, FLOPs, and the inference time for a single run of the model.


\subsection{Comparison to the State-of-the-art Methods}
To provide a comprehensive evaluation of our proposed model, we compare it against two aspects of related work: (i) existing gaze estimation models, and (ii) works related to lightweight variants of Vision Transformers (ViT).

\subsubsection{Gaze estimation comparison}
Our methodology is compared to established gaze estimation models\cite{zhang_its_2017,chen_appearance-based_2019,kellnhofer_gaze360_2019,oh_self-attention_2022,cheng_gaze_2022} by evaluating gaze error angles on the MPIIFaceGaze \cite{zhang_its_2017} and EYEDIAP\cite{funes_mora_eyediap_2014} datasets.


Our proposed FR-Net is evaluated using the same data pre-training methodology as the state-of-the-art (SOTA) work, GazeTR-Hybrid \cite{cheng_gaze_2022}. Specifically, we pre-trained our model on the ETH-XGaze dataset \cite{zhang_eth-xgaze_2020}, which consists of 1.1 million high-resolution images of 110 individuals with diverse ethnicities, head positions, and viewing orientations from multiple perspectives. For training, we used 765,000 normalized images of 80 subjects with a resolution of $224\times224$. To satisfy the input requirements of our model, we transformed these images into $256\times256$.

\begin{table}[h]
    \begin{center}
    \resizebox{0.45\textwidth}{!}{
    \begin{tabular}{c|c|c}
        \hline
            Methods & MPIIFaceGaze &  EYEDIAP \\
            \hline
            Spatial Weights CNN\cite{zhang_its_2017} & 4.93° & 6.53° \\
            Dilated-Net\cite{chen_appearance-based_2019} & 4.42° & 6.19° \\
            Gaze360\cite{kellnhofer_gaze360_2019} & 4.06° & 5.36° \\
            CA-Net\cite{cheng_coarse--fine_2020}  & 4.27° & 5.27° \\
            \cite{oh_self-attention_2022}  & 4.04° & 5.25° \\
            GazeTR-Hybrid\cite{cheng_gaze_2022} & 4.00° & 5.17° \\
            \textbf{FR-Net}& \textbf{3.86°} & \textbf{4.51°} \\
        \hline
    \end{tabular}}
    \end{center}
    \caption{Comparison with state-of-the-art gaze estimation methods in the angle error}
    \label{tab:gaze angular}
\end{table} 

Table \ref{tab:gaze angular} presents a comparison of our proposed FR-Net with existing methods in terms of gaze error, and our results demonstrate a 0.14° increase in minimum gaze error angle on MPIIFaceGaze (from 4° to 3.86°) and a 0.66° improvement in minimum gaze error angle on EYEDIAP (from 5.17° to 4.51°).

Performance results are depicted in Table \ref{tab:performance}. Our model results in a reduction of the number of parameters for gaze estimation to under one million, achieving a significant improvement of 0.67M parameters. It represents a reduction of $5-17$ times compared to the existing model. The optimization of FLOPs was achieved at 0.22 billion, significantly reducing resource consumption for mobile devices. In this aspect of inference time, our algorithm also provides a slight improvement over existing practices. It should be noted that the inference time is verified on the CPU, which is closer to the practical application.


\begin{table}[h]
    \begin{center}
    \resizebox{0.45\textwidth}{!}{
    \begin{tabular}{c|c|c|c}
        \hline
            Methods & \#Params(M) & \ FLOPs(B)  & Time(ms) \\
            \hline  
            Dilated-Net\cite{chen_appearance-based_2019} & 3.92 & 3.15 & 29 \\
            Gaze360\cite{kellnhofer_gaze360_2019} & 11.9 & 7.29 & 62 \\
            GazeTR-Hybrid\cite{cheng_gaze_2022} & 11.4 & 1.83 & 24 \\
            \textbf{FR-Net} & \textbf{0.67} & \textbf{0.22} & \textbf{23}\\
        \hline
    \end{tabular}}
    \end{center}
    \caption{Comparison with the state-of-the-art gaze estimation methods in Parameters, FLOPs and Inference Time}
    \label{tab:performance}
\end{table} 

The experimental results presented in Tables \ref{tab:gaze angular} and \ref{tab:performance} demonstrate the state-of-the-art performance of the FR-Net model on the MPIIFaceGaze and EYEDIAP datasets. Specifically, the approach not only achieves a reduction in gaze error angle, but also significantly decreases the number of model parameters and FLOPs, resulting in improved model performance. The experimental results demonstrate the proposed model could efficiently achieve the fusion extraction of frequency and time domain features, resulting in improved accuracy and efficiency.


\subsubsection{Comparison with ViT lightweight models}
In this study, we aimed to investigate the performance of a relatively lightweight ViT method for the gaze estimation task. To this end, we adopted 
the same training strategy as FR-Net and compare our approach with DeiT-S\cite{touvron_training_2021}, LeViT-128S \cite{graham_levit_2021} and T2T-ViT-7\cite{yuan_tokens--token_2021}.


\begin{table}[h]
    \begin{center}
    \resizebox{0.4\textwidth}{!}{
    \begin{tabular}{c|c|c}
        \hline
            Methods & MPIIFaceGaze &  EYEDIAP \\
            \hline
            DeiT-S\cite{touvron_training_2021} & 6.11° & 6.39°  \\
            LeViT-128S \cite{graham_levit_2021}& 5.33° & 5.62° \\
            T2T-ViT-7\cite{yuan_tokens--token_2021}& 4.99° & 5.54° \\
            \textbf{FR-Net}& \textbf{3.86°} & \textbf{4.51°} \\
        \hline
    \end{tabular}}
    \end{center}
    \caption{Comparison with ViT lightweight models in the error angle}
    \label{tab:gaze angular ViT}
\end{table} 




The performance of lightweight ViT models on MPIIFaceGaze and EYEDIAP datasets is suboptimal, as presented in Table \ref{tab:gaze angular ViT}. Facial images are inherently complex and contain implicit features related to gaze, including head pose, illumination, and resolution. The lightweight ViT approach efforts to optimize the number of parameters and FLOPs. Its inadequate performance in estimating gaze can be attributed to its limited capability to extract and simultaneously incorporate these intricate, invisible feature information.

\begin{table}[h]
    \begin{center}
    \resizebox{0.45\textwidth}{!}{
    \begin{tabular}{c|c|c|c}
        \hline
            Methods & \ Params(M) & \ FLOPs(B)  & Time(ms) \\
            \hline  
            DeiT-S\cite{touvron_training_2021} & 21.67 & 4.25 & 43 \\
            LeViT-128S \cite{graham_levit_2021}& 7.00 & 0.28 & 10 \\
            T2T-ViT-7\cite{yuan_tokens--token_2021}& 4.00 & 0.98 & 23 \\
            \textbf{FR-Net} & \textbf{0.67} & \textbf{0.22} & \textbf{23}\\
        \hline
    \end{tabular}}
    \end{center}
    \caption{Comparison with ViT lightweight methods in Parameters, FLOPs and Inference Time}
    \label{tab:performance ViT}
\end{table} 

Table. \ref{tab:performance ViT} illustrates the results. With respect to lightweight optimization, our model outperforms the lightweight ViT models in terms of parameter efficiency and exhibits a slight lead in terms of FLOPs and inference time. As a result, our model demonstrates superiority over the lightweight ViT model in both accuracy and efficiency for the task of gaze estimation.

\subsection{Visualization of learned features}
To determine whether our model can effectively extract eye-related features, we designed a visualization scheme for FFT Residual Block. We utilize the first FFT Residual Block to generate a feature map of the face image, which is then turned into a single-channel feature map. The information contained within the feature map is then visually represented. The visualization highlights the attention allocation of the model towards the input data, with regions of brighter color indicating a greater degree of attention, while darker regions reflect less attention. The presented results serve to illustrate the effectiveness of FFT Residual Block in extracting gaze-related features from facial images.

\begin{figure}[htbp]
\centering
\includegraphics[width=0.48 \textwidth]{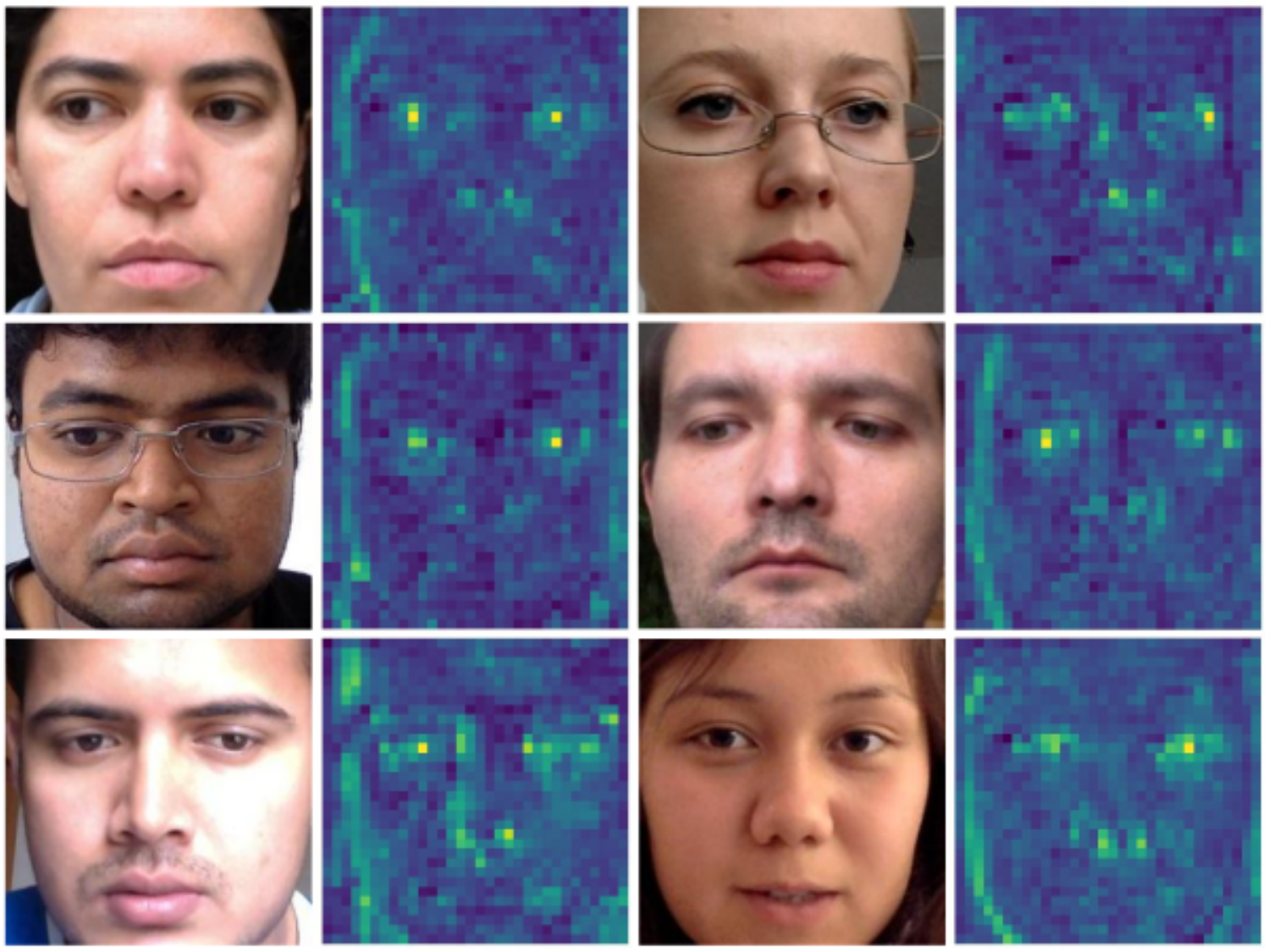}
\caption{Our proposed model effectively captures distinctive features of the eye region, as well as other visually-relevant characteristics, including head pose. This emphasizes the practicality of our approach in accurately estimating gaze patterns from facial images.}
\label{Fig:7 heatmap}
\end{figure}


In order to visually demonstrate the effectiveness of our FR-Net model in extracting pertinent information from facial images, we selected a diverse range of face images sourced from the MPIIFaceGaze dataset for feature map visualization, as presented in Figure \ref{Fig:7 heatmap}. The visualized feature map clearly highlights the eye region and facial contour as the most salient features, as evidenced by their notably brighter appearance. 

The findings indicate that our model is able to find face and eye features accurately and efficiently as expected.

\subsection{Ablation Study}
To better understand the effect of different parts on our FR-Net, we remove the FFT Residual Block, FFT Encoder, and shortcut component in FFT Encoder and concatenating shortcut  from the backbone model respectively.

The results, as presented in Table \ref{tab:ablation study}, indicate a substantial impact of FFT Residual Block on gaze accuracy. The error angle in the network without the FFT Residual Block increased by $0.75°$ and $0.96°$ on the MPIIFaceGaze and EYEDIAP datasets, respectively, underscoring the significance of the FFT Residual Block in the FR-Net model. For core part of FFT Residual Block, The FFT encoder, the core part of FFT Residual Block, shows an important role on improve the accuracy, specially on EYEDIAP. It even improves much more than FFT Residual Block. Up until now, we haven't found the reason. It provides us with clues for further improvement later on.  



\begin{table}[h]
    \begin{center}
    \resizebox{0.45\textwidth}{!}{
    \begin{tabular}{c|c|c}
        \hline
            Methods & MPIIFaceGaze &  EYEDIAP\\
            \hline
            FR-Net & 3.86°& 4.51° \\
            - FFT RB & 4.61° & 5.45° \\
            - FFT Encoder & 4.08° & 6.29°\\
            - Concatenation shortcut & 3.92° & 4.70° \\    
            - FFT Encoder shortcut & 3.88° & 4.82° \\
        \hline
    \end{tabular}}
    \end{center}
    \caption{Ablation study results for the error angle, Symbol ‘-’ indicates the following component is removed}
    \label{tab:ablation study}
\end{table} 

According to Section \ref{section_FRB}, we hypothesize that shortcuts are crucial, particularly in networks using large kernels, as demonstrated by prior work \cite{ding_scaling_2022}. To verify this hypothesis, we conducted ablation experiments on shortcut connections. Specifically, we removed the shortcut component in the FFT Encoder and shortcut concatenation outside of the FFT Encoder from the original model. The results, presented in Table \ref{tab:ablation study}, indicate that the shortcut inside the FFT Encoder has minimal impact on the model's performance, with similar results to the original model on the MPIIFaceGaze dataset (0.06° higher) and slightly improved performance on the EYEDIAP dataset (0.19° higher). Similarly, the gaze error angle was 0.06° and 0.19° higher than the original model after removing the shortcut concatenation outside of the FFT Encoder on the MPIIFaceGaze and EYEDIAP datasets, respectively. In conclusion, shortcut connections do have some impact, but not as significant as initially expected.



In summary, the results of the ablation study indicate that the FFT Resident Block can effectively extract the frequency domain features and that the shortcut connection with FFT can enhance the gaze-related factors and minimize the final gaze error angle.

\section{Limitation and future work}
Based on the experimental results presented in Section 4, our model demonstrates outstanding performance with respect to angle error, parameters, and FLOPs. However, the experimental findings presented herein demonstrate that the benefits of our model in terms of inference time are not as pronounced as its parameter and FLOP count would suggest. There exist multiple factors that affect the real inference time. Based on our analysis, we contend that the FFTs algorithm demonstrates an advantage in computing complexity. However, due to a lack of tight integration with the existing deep learning framework, such as convolution, it is susceptible to the effects of hardware and mate operators. Consequently, its advantage in real inference is not particularly notable.

In terms of the role of the different parts of the designed model, there are still some discrepancies observed from our prior design expectations. Despite extensive experimentation and analysis, a specific reason for these deviations remains elusive. Further investigation and exploration are necessary to fully comprehend and reconcile these differences.

In the future, to further enhance the applicability of our model across diverse computation-constrained devices, it is imperative to analyze and optimize related mate operators to maximize its efficacy.

\section{Conclusion}
This paper presents FR-Net, a novel lightweight model for gaze estimation that outperforms state-of-the-art models in terms of accuracy while maintaining an efficient structure. Our proposed approach leverages the power of frequency domain features via FFT Residual Block and MobileViT v3 to extract and enhance important information related to gaze. To achieve this, we introduce a trainable mask with low computation complexity that helps to focus on crucial information. Furthermore, we incorporate a shortcut component to the model to extract spatial features and improve accuracy.

We evaluated our approach on two widely used gaze estimation datasets, namely MPIIFaceGaze and EYEDIAP. The experimental results demonstrate that FR-Net achieves minimal error angle, outperforming state-of-the-art gaze estimation models while maintaining a lightweight structure. We also compared our model's performance with other lightweight ViT approaches, and our model showed competitive results. Our proposed approach offers a promising direction for improving gaze estimation in real-world scenarios, such as driver monitoring systems and human-computer interaction applications.

{\small
\bibliographystyle{ieee_fullname}
\bibliography{egbib}
}

\end{document}